\documentclass{article}
\usepackage{eurospeech_01,amssymb,amsmath,graphicx}
\ninept
\def\reg{{\rm\ooalign{\hfil
     \raise.07ex\hbox{\scriptsize R}\hfil\crcr\mathhexbox20D}}}

\title{A Cross-media Retrieval System for Lecture Videos}

\makeatletter
\def\name#1{\gdef\@name{#1\\}}
\makeatother
\name{{\em Atsushi Fujii$^{\dagger,\dagger\dagger\dagger}$, Katunobu Itou$^{\dagger\dagger,\dagger\dagger\dagger}$, Tomoyosi Akiba$^{\dagger\dagger}$, Tetsuya Ishikawa$^{\dagger}$}}
\address{$^{\dagger}$ Institute of Library and Information Science,
  University of Tsukuba \\
  $^{\dagger\dagger}$ National Institute of Advanced Industrial Science
  and Technology \\
  $^{\dagger\dagger\dagger}$ CREST, Japan Science and Technology
  Corporation}
\newcommand{\eq}[1]{(\ref{#1})}

\begin{document}
\maketitle
\begin{abstract}
  We propose a cross-media lecture-on-demand system, in which users
  can selectively view specific segments of lecture videos by
  submitting text queries. Users can easily formulate queries by
  using the textbook associated with a target lecture, even if they
  cannot come up with effective keywords. Our system extracts the
  audio track from a target lecture video, generates a transcription
  by large vocabulary continuous speech recognition, and produces a
  text index. Experimental results showed that by adapting speech
  recognition to the topic of the lecture, the recognition accuracy
  increased and the retrieval accuracy was comparable with that
  obtained by human transcription.
\end{abstract}

\section{Introduction}
\label{sec:introduction}

The growing number of multimedia contents available via the World Wide
Web, CD-ROMs, and DVDs has made information technologies incorporating
speech, image, and text processing crucial. Of the various types of
contents, lectures (audio/video) are typical and a valuable multimedia
resource, in which speeches (i.e., oral presentations) are usually
organized based on text materials, such as resumes, slides, and
textbooks. In lecture videos, image information, such as flip charts,
is often also used. In other words, a single lecture consists of
different types of compatible multimedia contents.

Because a single lecture often refers to several topics and takes a
long time, it is useful to obtain specific segments (passages)
selectively so that the audience can satisfy their information needs
at minimum cost. To resolve this problem, in this paper we propose a
lecture-on-demand system that retrieves relevant video/audio passages
in response to user queries. For this purpose, we utilize the benefits
of different media types to improve retrieval performance.

On the one hand, text has the advantage that users can view/scan the
entire contents quickly and can easily identify relevant passages
using the layout information (e.g., text structures based on sections
and paragraphs). In other words, text contents can be used for
random-access purposes.  On the other hand, speech is used mainly for
sequential-access purposes.  Therefore, it is difficult to identify
relevant passages unless target video/audio data includes additional
annotation, such as indexes.  Even if the target data are indexed,
users are not necessarily able to provide effective queries.  To
resolve this problem, textbooks are desirable materials from which
users can extract effective keywords and phrases.  However, while
textbooks are usually concise, speech has a high degree of redundancy
and therefore is easier to understand than textbooks, especially where
additional image information is provided.

In view of the above, we model our lecture-on-demand (LOD) system as
follows. A user selects text segments (keywords, phrases, sentences,
and paragraphs) that are relevant to their information needs from a
textbook for a target lecture.  By using selected segments, a text
query is generated automatically. That is, queries can be formulated
even if users cannot provide effective keywords.  Users can also
submit additional keywords as queries, if necessary.  Video passages
relevant to a given query are retrieved and presented to the user.  To
retrieve the video passages in response to text queries, we extract
the audio track from a lecture video, generate a transcription by
means of large vocabulary continuous speech recognition, and produce a
text index, prior to system use.  Our system is a cross-media
system in the sense that users can retrieve video and audio
information by means of text queries.

\section{System Description}
\label{sec:system}

\subsection{Overview}
\label{subsec:system_overview}

\begin{figure}[htbp]
  \begin{center}
    \leavevmode
    \includegraphics[height=2.9in]{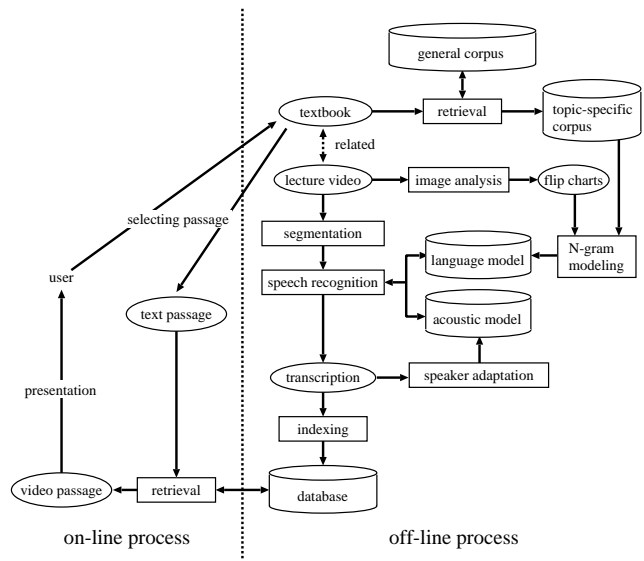}
  \end{center}
  \caption{An overview of our lecture-on-demand system.}
  \label{fig:system}
\end{figure}

Figure~\ref{fig:system} depicts the overall design of our
lecture-on-demand system, in which the left and right regions
correspond to the on-line and off-line processes, respectively.
Although our system is currently implemented for Japanese, our
methodology is fundamentally language independent.  For the purpose of
research and development, we tentatively target lecture programs on TV
for which textbooks are published. We explain the basis of our system
using Figure~\ref{fig:system}.

In the off-line process, given the video data of a target lecture,
audio data are extracted and segmented into a number of
passages. Then, a speech recognition system transcribes each
passage. Finally, the transcribed passages are indexed as in
conventional text retrieval systems, so that each passage can be
retrieved efficiently in response to text queries.  To adapt speech
recognition to a specific lecturer, we perform unsupervised speaker
adaptation using an initial speech recognition result (i.e., a
transcription).  To adapt speech recognition to a specific topic, we
perform language model adaptation, for which we search a general
corpus for documents relevant to the textbook related to a target
lecture. Then, retrieved documents (i.e., a topic-specific corpus) are
used to produce a word-based N-gram language model.  We also perform
image analysis to extract text (e.g., keywords and phrases) from flip
charts. These contents are also used to improve our language model.

In the on-line process, a user can view specific video passages by
submitting any text queries, i.e., keywords, phrases, sentences, and
paragraphs, extracted from the textbook. Any queries not in the
textbook can also be used.  The current implementation is based on a
client-server system on the Web. Both the off-line and on-line
processes are performed on servers, but users can access our system
using Web browsers on their own PCs.

\begin{figure}[htbp]
  \begin{center}
   \leavevmode
    \includegraphics[height=2.3in]{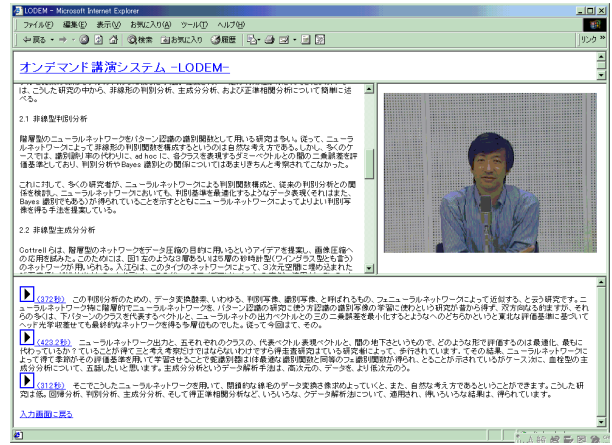}
  \end{center}
   \caption{The interface of our LOD system over the Web.}
   \label{fig:lodem}
\end{figure}

Figure~\ref{fig:lodem} depicts a prototype interface of our LOD
system, in which a lecture associated with ``nonlinear multivariate
analysis'' is given.  In this interface,  an electronic version of a
textbook is displayed on the left side, and a lecture video is played
on the right side. In addition, users can submit any text queries in
the input box, which is not depicted in Figure~\ref{fig:lodem}. In
this scenario, a text paragraph related to ``discriminant analysis''
was copied and pasted into the query input box, and top-ranked
transcribed passages for the query were listed according to the degree
of relevance (in the lower part of Figure~\ref{fig:lodem}). Users can
select (click on) transcriptions to play the corresponding video
passage.

It should be noted that unlike conventional keyword-based retrieval
systems, in which users usually submit a small number of keywords, in
our system users can easily submit longer queries using textbooks.
Where submitted keywords are misrecognized in transcriptions, the
retrieval accuracy decreases.  However, longer queries are relatively
robust for speech recognition errors, because the effect of
misrecognized words is overshadowed by the large number of words
correctly recognized.

\subsection{Passage Segmentation}
\label{subsec:passage}

The basis of passage segmentation is to divide the entire video data
for a single lecture into more than one unit to be retrieved.  We call
these smaller units ``passages''.  For this purpose, both speech and
image data can provide promising clues. However, in lecture TV
programs, it is often the case that a lecturer sitting still is the
main focus and a small number of flip charts are used occasionally. In
such cases, image data is less informative.  Therefore, tentatively we
use only speech data for the passage segmentation process.  However,
segmentation can potentially vary depending on the user query. Thus,
it is difficult to predetermine a desirable segmentation in the
off-line process.

Because of the above problems, we first extract the audio track from a
target video and use a simple pause-based segmentation method to
obtain minimal speech units, such as sentences and long phrases. In
other words, speech units are continuous audio segments that do not
include pauses longer than a certain threshold. Finally, we generate
variable-length passages from one or more speech units. To put it more
precisely, we combine $N$ speech units into a single passage, with $N$
ranging from 1 to 5 in the current implementation.

\subsection{Speech Recognition}
\label{subsec:speech_recognition}

The speech recognition module generates word sequence $W$, given phone
sequence $X$. In a stochastic framework, the task is to select the $W$
maximizing $P(W|X)$, which is transformed as in Equation~\eq{eq:bayes}
through the Bayesian theorem.
\begin{equation}
  \label{eq:bayes}
  \arg\max_{W}P(W|X) = \arg\max_{W}P(X|W)\cdot P(W)
\end{equation}
$P(X|W)$ models the probability that the word sequence $W$ is
transformed into the phone sequence $X$, and $P(W)$ models the
probability that $W$ is linguistically acceptable. These factors are
called the acoustic and language models, respectively.

We use the Japanese dictation
toolkit\footnote{http://winnie.kuis.kyoto-u.ac.jp/dictation/}, which
includes the Julius decoder and acoustic/language models. Julius
performs a two-pass (forward-backward) search using word-based forward
bigrams and backward trigrams.  The acoustic model was produced from
the ASJ speech database, which contains approximately 20,000 sentences
uttered by 132 speakers including both gender groups. A 16-mixture
Gaussian distribution triphone Hidden Markov Model, in which states
are clustered into 2,000 groups by a state-tying method, is used.  We
adapt the provided acoustic model by means of an MLLR-based
unsupervised speaker adaptation method, for which in practice we use
the HTK toolkit\footnote{http://htk.eng.cam.ac.uk/}.

Existing methods to adapt language models can be classified into two
fundamental categories. In the first category---the {\em
  integration\/} approach---general and topic-specific corpora are
integrated to produce a topic-specific language
model~\cite{auzanne:riao-2000,seymore:eurospeech-97}. Because the
sizes of those corpora differ, N-gram statistics are calculated using
the weighted average of the statistics extracted independently from
those corpora. However, it is difficult to determine the optimal
weight depending on the topic.  In the second category---the {\em
  selection\/} approach---a topic-specific subset is selected from a
general corpus and is used to produce a language model. This approach
is effective if general corpora contain documents associated with
target topics, but N-gram statistics in those documents are
overshadowed by other documents in resultant language models.

We followed the selection approach, because the 10M Web page
corpus~\cite{eguchi:sigir-2002} containing mainly Japanese pages
associated with various topics was publicly available. The quality of
the selection approach depends on the method of selecting
topic-specific subsets. An existing
method~\cite{chen:adaptation_ws-2001} uses hypotheses in the initial
speech recognition phase as queries to retrieve topic-specific
documents from a general corpus. However, errors in the initial
hypotheses have the potential to decrease the retrieval
accuracy. Instead, we use textbooks related to target lectures as
queries to improve the retrieval accuracy and consequently the quality
of the language model adaptation.

\subsection{Retrieval}
\label{subsec:retrieval}

Given transcribed passages and text queries, the basis of the
retrieval module is the same as that for text retrieval.  We use an
existing probabilistic text retrieval method~\cite{robertson:sigir-94}
to compute the relevance score between the query and each passage in
the database.  The relevance score for passage $p$ is computed by
Equation~\eq{eq:okapi}.
\begin{equation}
  \footnotesize
  \label{eq:okapi}
  \sum_{t} f_{t,q}\cdot\frac{\textstyle (K+1)\cdot
  f_{t,p}}{K\cdot\{(1-b)+\textstyle\frac{\textstyle
  dl_{p}}{\textstyle b\cdot avgdl}\} +
  f_{t,p}}\cdot\log\frac{\textstyle N - n_{t} + 0.5}{\textstyle
  n_{t} + 0.5}
\end{equation}
where $f_{t,q}$ and $f_{t,p}$ denote the frequency with which term $t$
appears in query $q$ and passage $p$, respectively. $N$ and $n_{t}$
denote the total number of passages in the database and the number of
passages containing term $t$, respectively. $dl_{p}$ denotes the
length of passage $p$, and $avgdl$ denotes the average length of
passages in the database. We empirically set $K=2.0$ and $b=0.8$,
respectively.  We use content words, such as nouns, extracted from
transcribed passages as index terms, and perform word-based
indexing. We use the ChaSen morphological
analyzer\footnote{http://chasen.aist-nara.ac.jp/} to extract content
words. The same method is used to extract terms from queries.

However, retrieved passages are not disjoint, because top-ranked
passages often overlap with one another in terms of the temporal
axis. It is redundant simply to list the top-ranked retrieved passages
as they are.  Therefore, we reorganize those overlapped passages into
a single passage.  The relevance score for a group (a new passage) is
computed by averaging the scores of all passages belonging to the
group. New passages are sorted according to the degree of relevance
and are presented to users as the final result.

\section{Experimentation}
\label{sec:experimentation}

\subsection{Methodology}
\label{subsec:ex_method}

To evaluate the performance of our LOD system, we produced a test
collection (as a benchmark data set) and performed experiments
partially resembling a task performed in the TREC spoken document
retrieval (SDR) track~\cite{garofolo:trec-97}.  Five lecture programs
on TV (each lecture was 45 minutes long), for which printed textbooks
were also published, were videotaped in DV and were used as target
lectures. Each lecture was manually transcribed and sentence
boundaries with temporal information (i.e., correct speech units) were
also identified manually.
Each paragraph in the corresponding textbook was used as a query
independently. For each query, a human assessor (a graduate student
not an author of this paper) identified one or more relevant sentences
in the human transcription.

Using our test collection, we evaluated the accuracy of speech
recognition and passage retrieval.
For the five lectures, our system used the sentence boundaries in
human transcriptions to identify speech units, and performed speech
recognition. We also used human transcriptions as perfect speech
recognition results and investigated the extent to which speech
recognition errors affect the retrieval accuracy.  Our system
retrieved top-ranked passages in response to each query.  Note that
the passages here are those grouped based on the temporal axis, which
should not be confused with those obtained from the passage
segmentation method.

\subsection{Results}
\label{subsec:results}

\begin{table*}[htbp]
  \begin{center}
     \caption{Experimental results for speech recognition and passage
       retrieval.}
    \medskip
    \leavevmode
    \footnotesize
    \begin{tabular}{llccccccccccccccc} \hline\hline
      ID &
      & \multicolumn{3}{c}{\#1}
      & \multicolumn{3}{c}{\#2}
      & \multicolumn{3}{c}{\#3}
      & \multicolumn{3}{c}{\#4}
      & \multicolumn{3}{c}{\#5}
      \\
      \hline
      Topic & &
      \multicolumn{3}{c}{Criminal law} &
      \multicolumn{3}{c}{Greek history} &
      \multicolumn{3}{c}{Domestic relations} &
      \multicolumn{3}{c}{Food and body} &
      \multicolumn{3}{c}{Solar system}
      \\
      \hline
      &
      & HUM &
      {\hfill\centering ASR\hfill} &
      {\hfill\centering +LA\hfill}
      & HUM &
      {\hfill\centering ASR\hfill} &
      {\hfill\centering +LA\hfill}
      & HUM &
      {\hfill\centering ASR\hfill} &
      {\hfill\centering +LA\hfill}
      & HUM &
      {\hfill\centering ASR\hfill} &
      {\hfill\centering +LA\hfill}
      & HUM &
      {\hfill\centering ASR\hfill} &
      {\hfill\centering +LA\hfill}
      \\
      \hline
      \multicolumn{2}{l}{OOV}
      & {\hfill\centering ---\hfill} & .044 & .020
      & {\hfill\centering ---\hfill} & .073 & .082
      & {\hfill\centering ---\hfill} & .039 & .049
      & {\hfill\centering ---\hfill} & .053 & .041
      & {\hfill\centering ---\hfill} & .051 & .053
      \\
      \multicolumn{2}{l}{PP}
      & {\hfill\centering ---\hfill} & 48.9 & 43.2
      & {\hfill\centering ---\hfill} & 122 & 96.7
      & {\hfill\centering ---\hfill} & 136 & 132
      & {\hfill\centering ---\hfill} & 89.3 & 108
      & {\hfill\centering ---\hfill} & 163 & 130
      \\
      \multicolumn{2}{l}{WER}
      & {\hfill\centering ---\hfill} & .209 & .133
      & {\hfill\centering ---\hfill} & .516 & .423
      & {\hfill\centering ---\hfill} & .604 & .543
      & {\hfill\centering ---\hfill} & .488 & .416
      & {\hfill\centering ---\hfill} & .637 & .482
      \\
      \hline
        & R & .695 & .726 &.732 & .449 & .258 & .551
        & .632 & .291 & .505 & .451 & .220 & .357 & .296 & .138 & .241 \\
        $N$=1 & P & .534 & .548 & .519 & .377 & .319 & .386
        & .479 & .362 & .464 & .414 & .277 & .337 & .529 & .358 & .436 \\
        & F & .604 & .624 & .607 & .410 & .286 & .454
        & .545 & .322 & .484 & .432 & .245 & .347 & .379 & .200 & .311 \\
      \hline
        & R & .847 & .858 & .832 & .663 & .360 & .674
        & .791 & .464 & .677 & .655 & .380 & .463 & .482 & .228 & .421 \\
        $N$=2 & P & .441 & .448 & .458 & .301 & .211 & .314
        & .372 & .273 & .353 & .321 & .247 & .239 & .462 & .332 & .409 \\
        & F & .580 & .588 & .591 & .414 & .266 & .429
        & .506 & .343 & .464 & .431 & .300 & .316 & .472 & .270 & .415 \\
      \hline
        & R & .879 & .868 & .874 & .764 & .438 & .708
        & .827 & .495 & .718 & .718 & .392 & .604 & .637 & .289 & .527 \\
        $N$=3 & P & .410 & .405 & .401 & .269 & .163 & .252
        & .363 & .215 & .318 & .297 & .188 & .235 & .466 & .280 & .385 \\
        & F & .560 & .553 & .550 & .398 & .237 & .372
        & .505 & .300 & .441 & .420 & .254 & .338 & .538 & .285 & .445 \\
      \hline
    \end{tabular}
    \label{tab:results}
  \end{center}
\end{table*}

To evaluate the accuracy of speech recognition, we used the word error
rate (WER), which is the ratio of the number of word errors (deletion,
insertion, and substitution) to the total number of words. We also
used test-set out-of-vocabulary rate (OOV) and trigram test-set
perplexity (PP) to evaluate the extent to which our language model
adapted to the target topics.  We used human transcriptions as test
set data.  For example, OOV is the ratio of the number of word tokens
not contained in the language model for speech recognition to the
total number of word tokens in the transcription. Note that smaller
values of OOV, PP, and WER are obtained with better methods.

The final outputs (i.e., retrieved passages) were evaluated based on
recall and precision, averaged over all queries. Recall (R) is the
ratio of the number of correct speech units retrieved by our system to
the total number of correct speech units for the query in question.
Precision (P) is the ratio of the number of correct speech units
retrieved by our system to the total number of speech units retrieved
by our system. To summarize recall and precision into a single
measure, we used the F-measure (F).

Table~\ref{tab:results} shows the accuracy of speech recognition (WER)
and passage retrieval (R, P, and F), for each lecture. In this table,
the columns ``HUM'' and ``ASR'' correspond to the results obtained
with human transcriptions and automatic speech recognition,
respectively. The column ``+LA'' denotes results for ASR combined with
language model adaptation.  The column ``Topic'' denotes topics for
the five lectures.

To adapt language models, we used the textbook corresponding to a
target lecture and searched the 10M Web page corpus for 2,000 relevant
pages, which were used as a source corpus. In the case where the
language model adaptation was not performed, all 10M Web pages were
used as a source corpus. In either case, 20,000 high frequency words
were selected from a source corpus to produce a word-based trigram
language model. We used the ChaSen morphological analyzer to extract
words (morphemes) from the source corpora, because Japanese sentences
lack lexical segmentation.

In passage retrieval, we regarded the top $N$ passages as the final
outputs. In Table~\ref{tab:results}, the value of $N$ ranges from 1 to
3. As the value of $N$ increases, the recall improves, but potentially
sacrificing precision.

\subsection{Discussion}
\label{subsec:discussion}

By comparing the results of ASR and +LA in Table~\ref{tab:results},
for some cases OOV and PP increased by adapting language models.
However, WER decreased by adapting language models to target topics,
irrespective of the lecture.

The values of OOV, PP, and WER for lecture~\#1 were generally smaller
than those for the other lectures. One possible reason is that the
lecturer of \#1 spoke more fluently and made fewer erroneous
utterances than the other lecturers.

Recall, precision, and F-measure increased by adapting language models
for lectures~\#2-5, irrespective of the number of passages retrieved.
For lecture~\#1, the retrieval accuracy did not significantly differ
whether or not we adapted the language model to the topic. One
possible reason is that the WER of lecture~\#1 without language model
adaptation (20.9\%) was sufficiently small to obtain a retrieval
accuracy comparable with the text retrieval~\cite{jourlin:sc-2000}.
The difference between HUM and ASR was marginal in terms of the
retrieval accuracy. Therefore, the effect of the language model
adaptation method was overshadowed in passage retrieval.

The retrieval accuracy for lecture~\#1 was higher than those for the
other lectures. The story of lecture~\#1 was organized based primarily
on the textbook, when compared with the other lectures. This suggests
that the performance of our LOD system is dependent of the
organization of target lectures.

Surprisingly, for lectures~\#1 and \#2, recall, precision, and
F-measure of +LA were better than those of HUM. This means that the
automatic transcription was more effective than human transcription
for passage retrieval purposes.  One possible reason is  the existence
of Japanese variants (i.e., more than one spelling form corresponding
to the same word), such as ``{\it girisha\/}/{\it
  girishia\/}~(Greece)''. Because the language model was adapted by
means of the textbook for a target lecture, the spelling in automatic
transcriptions systematically resembled that in the queries extracted
from the textbooks. In contrast, it is difficult to standardize the
spelling in human transcriptions. Therefore, relevant passages in
automatic transcriptions were more likely to be retrieved than
passages in the human transcriptions.

We conclude that our language model adaptation method was effective
for both speech recognition and passage retrieval.

\section{Conclusion}
\label{sec:conclusion}

Reflecting the rapid growth in the use of multimedia contents,
information technologies appropriate to speech, image, and text
processing are crucial. Of the various content types in this paper we
focused on the video data of lectures with their organization based on
textbooks, and proposed a system for cross-media on-demand lectures,
in which users can formulate text queries using the textbook for a
target lecture to retrieve specific video passages.

To retrieve video passages in response to text queries, we extract the
audio track from a lecture video, generate a transcription by large
vocabulary continuous speech recognition, and produce a text index,
prior to system use.

We evaluated the performance of our system experimentally, for which
five TV lecture programs in various topics were used. The experimental
results showed that the accuracy of speech recognition varied
depending on the topic and presentation style of the
lecturers. However, the accuracy of speech recognition and passage
retrieval was improved by adapting language models to the topic of the
target lecture. Even if the word error rate was approximately 40\%,
the accuracy of retrieval was comparable with that obtained by human
transcription.

\bibliographystyle{IEEEtran}

\end{document}